%% file: arxiv_2026.tex
\documentclass{article}

\usepackage[preprint]{neurips_2026}

\usepackage[utf8]{inputenc}
\usepackage[T1]{fontenc}
\usepackage{hyperref}
\usepackage{url}
\usepackage{booktabs}
\usepackage{amsfonts}
\usepackage{amsmath}
\usepackage{nicefrac}
\usepackage{microtype}
\usepackage{xcolor}
\usepackage{graphicx}
\usepackage{siunitx}
\usepackage{wrapfig}
\usepackage{multirow}

\newif\ifrevhighlight
\revhighlighttrue

\ifrevhighlight
  
\else
  
\fi

\definecolor{lineblue}{RGB}{20, 100, 190}
\definecolor{lineorange}{RGB}{250, 140, 60}
\definecolor{captionorange}{RGB}{200, 100, 0}
\definecolor{captionblue}{RGB}{0, 80, 160}

\definecolor{lineblue}{RGB}{20, 100, 190}
\definecolor{lineorange}{RGB}{250, 140, 60}

\title{Characterizing Job Power Elasticity for Power-Flexible AI Training}

\author{
  Philip Colangelo, Charles Dawson,
  \textbf{Shayan Sengupta, Ayse Coskun, Varun Sivaram} \\
  Emerald AI \\
}

\begin{document}

\maketitle

\begin{abstract}



Large language model (LLM) training is among the fastest-growing sources of electricity demand in modern data centers, and power availability is a primary bottleneck to continued AI infrastructure growth. Making the power consumption of these workloads flexible could unlock additional power for AI growth, limit increases in electricity prices, and improve the utilization of existing grid infrastructure. However, to realize this flexibility, we must first understand how the performance of training workloads changes when GPU power is reduced.


This paper presents the first systematic characterization of \emph{job power elasticity} (the sensitivity of throughput to power reductions) in LLM training.
To quantify elasticity, we introduce the \emph{Power Flexibility Index (PFI)}, a normalized metric that quantifies the performance cost of power reductions and provides a control primitive for SLA-aware power flexibility.


We collect data from 131 LLM training runs on H200 (plus 24 H200 validation runs and 34 matched H100 runs), including both dense and mixture-of-experts models, pretraining and fine-tuning tasks, and up to 32 GPUs. We find that LLM training jobs exhibit substantial but variable power elasticity, and we identify telemetry signals that predict PFI at runtime. Finally, we demonstrate that PFI-aware power allocation maximizes total tokens/second throughput under power constraints. Under a 30\% power reduction, PFI-aware power allocation recovers $\sim$1.5k tokens/s per job, 63\% of the performance gap between an equal-weight allocation and an oracle with perfect information.
Our results establish power elasticity as a measurable property of training jobs and provide a foundation for power-aware, grid-responsive AI infrastructure.
\end{abstract}

\section{Introduction}
\input{intro}

\section{Power Elasticity and the Power Flexibility Index}


A job's power elasticity describes the sensitivity of its throughput to sustained reductions in available power. Highly elastic jobs experience rapid throughput degradation under GPU power capping, whereas inelastic jobs preserve throughput. Figure~\ref{fig:hero} illustrates representative elastic and inelastic power–throughput regimes.

Power elasticity is important because it enables throughput-aware power capping, allowing cluster power managers to reduce GPU power consumption while minimizing throughput loss.
Despite its importance, two gaps remain in the current literature. First, there is no accepted quantitative definition of power elasticity that permits systematic comparison across jobs, model architectures, and hardware settings. Second, there is limited understanding of the job characteristics that determine elasticity or enable its prediction \textit{a priori}.

To close these gaps, we begin by developing a formal definition of a \textbf{power flexibility index (PFI)}\footnote{Note that under this definition a high-PFI job has performance that is relatively inelastic with respect to power reduction and therefore more power-flexible.}.
Let $(T, P)_i$ be tuples of the average throughput (in tokens/s) and aggregate GPU power (in \si{\watt}) of a job measured under a range of per-GPU power caps $\bar{P}_i$, $i=1,\ldots,N$.\footnote{Throughout this paper, a \emph{job} refers to the combination of training task, model, GPU count, parallelism strategy, sequence length, and batch size; each job is run at multiple power caps to measure PFI. A \emph{workload} refers to a collection of jobs.}
$P_i$ is the sum of per-device GPU power readings across all GPUs participating in the job.
The realized $P_i$ generally falls below $G\cdot\bar{P}_i$ (where $G$ is the GPU count), since the cap is an upper bound and the job need not saturate.
We define $(T_{\mathrm{max}}, P_{\mathrm{max}})$ as the throughput/power pair measured without a power cap (i.e.\ with $\bar{P}_N$ set to thermal design power, TDP).

We define the power flexibility index as the ratio of power decrease to throughput decrease, averaged over measured power caps (excluding the uncapped measurement where $\Delta T=0$):
\begin{equation}
    PFI = \frac{1}{N}\sum_{i=1}^N \frac{\Delta P_i}{\Delta T_i} = \frac{1}{N}\sum_{i=1; T_i \neq T_{\rm{max}}}^N \frac{1 - P_i/P_{\mathrm{max}}}{1 - T_i/T_{\mathrm{max}}} \label{eq:pfi}
\end{equation}

While elasticity describes throughput sensitivity to power loss, PFI inverts this relationship into a control-oriented measure of how efficiently a job can shed power. Consequently, jobs with high elasticity exhibit low PFI, whereas throughput-preserving inelastic jobs exhibit high PFI.

PFI is a hardware-dependent property of a job: the formula normalizes for per-platform reference points $(T_{\mathrm{max}}, P_{\mathrm{max}})$ and so is comparable across models on the same accelerator, but values from different accelerator generations should not be compared directly because they reflect each platform's specific power-cap mechanism (§\ref{sec:discussion}). The metric has a straightforward interpretation:

    \textit{Inelastic ($PFI > 1$):} throughput declines more slowly than power (e.g., $-5\%$ power, $-2\%$ throughput).
    
    \textit{Linear ($PFI = 1$):} throughput is proportional to power (e.g., $-5\%$ power, $-5\%$ throughput).
    
    \textit{Elastic ($PFI < 1$):} throughput declines faster than power (e.g., $-5\%$ power, $-10\%$ throughput).

\textbf{Comparison to existing metrics.}
The closest existing metrics fall into two families. \textit{ML training efficiency metrics} such as MFU~\cite{chowdhery2023palm} and tokens/joule characterize steady-state efficiency at a fixed operating point; they carry no information about how throughput responds to power reduction. Zeus~\cite{you2023zeus} and Perseus~\cite{chung2024perseus} consider the energy--time Pareto frontier for individual jobs, but produce per-job
optimization outputs rather than a scalar characterization metric comparable across architectures and training tasks.
\textit{Power systems flexibility metrics} quantify load reduction capacity (MW), ramp rate, and response latency~\cite{eprimosaic} but ignore quality-of-service degradation. PFI unifies both perspectives: it is computable from external power measurements without job internals, normalized for cross-architecture comparison, and explicitly encodes the performance cost of power reduction.

\textbf{Driving questions.}
Given this definition, three questions motivate the rest of this paper:

\textbf{What factors influence power elasticity?} Does PFI vary across architectures (e.g. dense models vs. mixture-of-experts) or training tasks (e.g. pretraining vs. fine-tuning)? How does training scale affect PFI? Are there underlying mechanisms (e.g. memory bottlenecks) that explain these variations?

\textbf{Can we predict power elasticity in production workloads?} The definition of PFI in~\eqref{eq:pfi} requires measurements across multiple power caps, which limits its ability to predict job elasticity in production. Predicting PFI from readily available telemetry (e.g. NVIDIA DCGM~\cite{dcgm}) would allow ML engineers and data center operators to track power elasticity in real-time.

\textbf{Can we use PFI for real-time control and power-aware workload orchestration?} Prior works have demonstrated the ability of GPU clusters to respond to requests from grid operators~\cite{williams2026powerflex, colangelo2025grid}. Data center operators could use real-time PFI estimates to respond efficiently, allocating power reductions to jobs that can best tolerate power reductions and minimizing disruption to their tenants.

\section{Experimental Setup}\label{sec:setup}
\vspace{-0.1in}

To characterize power elasticity for representative LLM training jobs, we sweep per-GPU power caps from 200\,W to 700\,W (TDP), using 100\,W increments for 8-GPU jobs and a four-point sweep for 16- and 32-GPU jobs,
while pretraining and fine-tuning four open-weight LLMs on clusters of 8/16/32 GPUs, yielding 25 headline sweep groups (131 individual training runs across power caps) plus 4 held-out validation sweeps (24 runs). Complete data tables showing the number of runs for each model are included in Appendix~\ref{appendix:data_tables}.
All runs execute on Google Kubernetes Engine (GKE) \texttt{a3-ultragpu-8g} nodes (8$\times$ NVIDIA H200, 141\,GiB
HBM3e, 700\,W TDP), with multi-node jobs scheduled through Kueue. Clusters are
provisioned from a single Terraform module and replicated across six GCP regions; full infrastructure details are given in Appendix~\ref{appendix:detailed-setup}.

\begin{table}[tb]
  \centering
  \caption{\small{Experiment matrix. Each sweep group fixes one full configuration per job and varies only the per-GPU power cap; PFI is computed within a sweep group. 8-GPU groups sweep power caps in 100\,W increments from 200--700~W, 16- and 32-GPU groups sweep 200/500/600/700~W only. EP~$=$~expert parallelism degree
  (pretraining only); PT~$=$~pretraining; LoRA~$=$~rank-32 LoRA SFT. GC~$=$~gradient checkpointing.}}
  \label{tab:experiment_matrix}
  \resizebox{\textwidth}{!}{%
  \begin{tabular}{@{}llllllll@{}}
    \toprule
    Model & Training Task & GPUs & Seq\,Len & Batch & EP & Compile & GC \\
    \midrule
    gpt-oss-20b   & PT, LoRA & 8       & 2048, 4096       & 2, 8   & 1, 4, 8 & Y, N & Y, N \\
    Qwen3-30B-A3B & PT, LoRA & 8, 16   & 4096             & 2, 16  & 1, 8    & Y    & Y, N \\
    Qwen3-32B     & PT, LoRA & 8, 16, 32 & 2048, 4096, 8192 & 4, 8, 16 & --- & Y, N & Y \\
    Llama-3.1-70B & PT, LoRA & 8, 16   & 2048, 4096       & 2, 16  & ---     & Y, N & Y   \\
    \bottomrule
  \end{tabular}}
\end{table}

\begin{table}[tb]
  \centering
  \caption{\small{Models used in experiments, split into training and held-out validation sets. \emph{Active} params denote the per-token activation count for sparse Mixture-of-Experts (MoE) models.}}
  \label{tab:model_zoo}
  \small
    \begin{tabular}{@{}cllrrl@{}}
        \toprule
        & Model & Architecture & Total & Active & Routing  \\
        \midrule
        \multirow{4}{*}{\rotatebox[origin=c]{90}{\textit{\scriptsize Training}}} &
        gpt-oss-20b      & MoE decoder   & 21\,B   & $\approx$3.6\,B & 4 of 32 experts   \\
        & Qwen3-30B-A3B  & MoE decoder   & 30\,B   & $\approx$3\,B   & 8 of 128 experts  \\
        & Qwen3-32B      & Dense decoder & 32.8\,B & 32.8\,B         & ---               \\
        & Llama-3.1-70B  & Dense decoder & 70.6\,B & 70.6\,B         & ---               \\
        \midrule
        \multirow{3}{*}{\rotatebox[origin=c]{90}{\textit{\scriptsize Held out}}} &
        DeepSeek-V2-Lite             & MoE decoder   & 15.7\,B & $\approx$2.4\,B & 6+2 of 64 experts \\
        & Mistral-Small-24B-Base-2501 & Dense decoder & 24\,B   & 24\,B           & ---               \\
        & Gemma-4-26B-A4B             & MoE decoder   & 26\,B   & $\approx$4\,B   & 8 of 128 experts  \\
        \bottomrule
    \end{tabular}
\end{table}


\begin{wrapfigure}{r}{0.5\textwidth}
  \centering
  \vspace{-0.2in}
  \includegraphics[width=\linewidth]{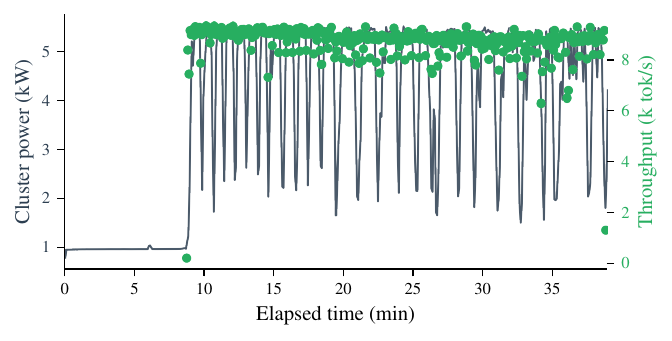}
  \caption{%
    \small{Aggregate GPU power and training throughput for a representative
    \textsc{Qwen3-32B} pretraining run on $32\times$H200 at \SI{700}{\watt}/GPU.}}
  \label{fig:experiment_time_series}
\end{wrapfigure}

We profile four open-weights LLMs spanning dense and Mixture-of-Experts (MoE)
architectures (\texttt{gpt-oss-20b}, \texttt{Qwen3-30B-A3B}, \texttt{Qwen3-32B}, and \texttt{Llama-3.1-70B}) under pretraining and rank-32 LoRA fine-tuning (Tables~\ref{tab:experiment_matrix},~\ref{tab:model_zoo}).
We chose sequence length and batch size to saturate GPU utilization at each cluster size, with gradient checkpointing applied where needed to fit activations in memory and \texttt{torch.compile} enabled where supported. We also profile three validation models and exclude their results from feature selection and predictor fitting: \texttt{DeepSeek-V2-Lite} for MoE pretraining, \texttt{Mistral-Small-24B-Base-2501} for dense pretraining and dense LoRA SFT, and \texttt{Gemma-4-26B-A4B} for MoE LoRA SFT.

We enforce per-GPU power caps via \texttt{nvidia-smi -pl} before any CUDA context
allocation and verify using DCGM power limit counters. We collect GPU utilization,
power, and memory at 5\,s intervals via a DCGM sidecar; training
throughput and MFU are logged in each training step via a custom callback.
We partition each trace into warmup, startup, and training phases; all reported
metrics aggregate over the training phase only, using the mean of the top 10\% of
samples to exclude evals and checkpoints. A representative run is shown in
Fig.~\ref{fig:experiment_time_series}.

\begin{figure}[b]
  \centering
  \includegraphics[width=0.9\textwidth]{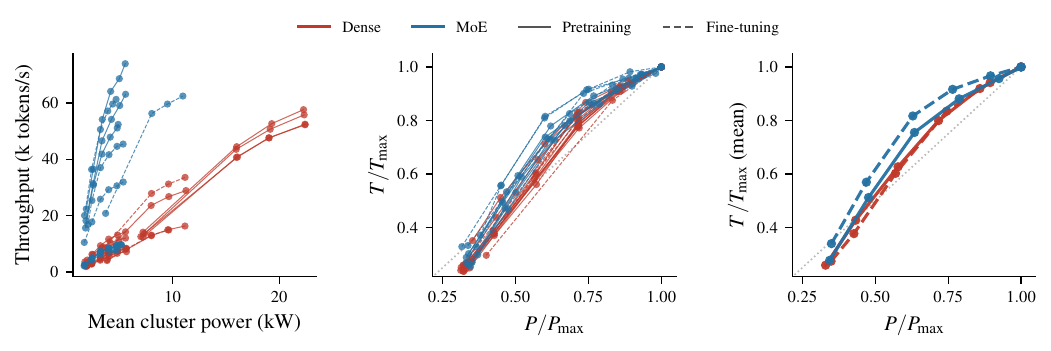}
  \caption{%
\small{    Power--performance curves for all sweep groups.
    \textbf{Left:} Aggregate GPU power vs.\ throughput.
    \textbf{Center:} Per-sweep normalized power vs.\ throughput.
    \textbf{Right:} Mean normalized power vs.\ throughput for each category (dense vs. MoE, pretraining vs LoRA SFT). Full data are provided in Appendix~\ref{appendix:data_tables}.}}
  \label{fig:power_performance}
\end{figure}

\begin{figure}[bh!]
  \centering
  \includegraphics[width=0.66\textwidth]{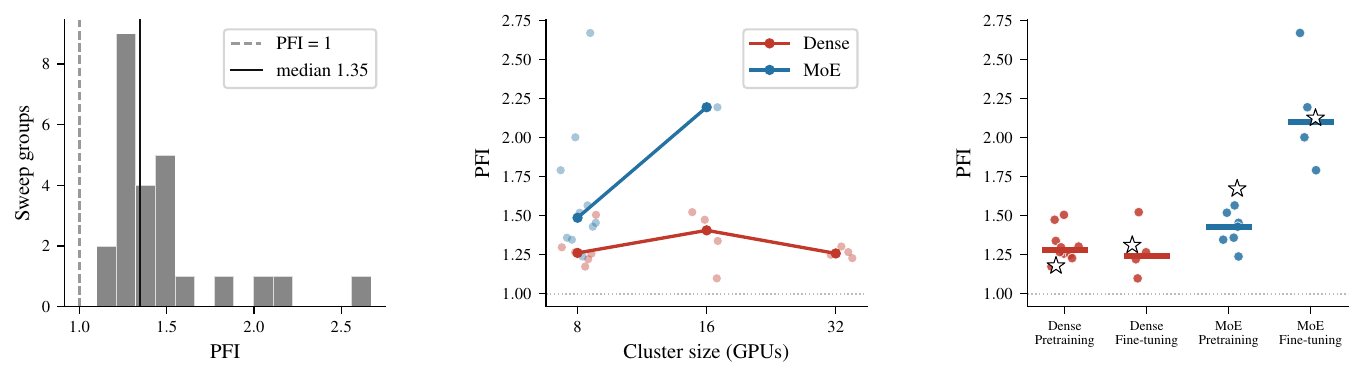}
  \caption{%
  \small{
    \textbf{Left:} PFI distribution across all jobs.
    \textbf{Right:} PFI by architecture and training task; median shown as horizontal bar.}}
  \label{fig:pfi}
\end{figure}


\section{Results}
\label{sec:results}

Three observations emerge from our study: PFI varies substantially across jobs, the variation depends on architecture and training task, and memory- and compute-related telemetry demonstrate strong correlations with PFI. Complete data tables are included in Appendix~\ref{appendix:data_tables}.


\textbf{PFI varies substantially across modern LLM training jobs.}
Fig.~\ref{fig:power_performance} shows the raw and normalized power--throughput curves for all runs in our dataset, and Fig.~\ref{fig:pfi} summarizes PFI across cluster sizes, model architectures, and training tasks. Across the cohort, we observe substantial variation in PFI (1.10--2.67), indicating that modern LLM training jobs differ in how much throughput they preserve under GPU power capping. Some jobs exhibit nearly linear power--throughput degradation, while others maintain high throughput under aggressive power reduction. We did not observe any jobs with PFI < 1; however, as we discuss in \S\ref{sec:exploit}, differences in relative PFI are still important for runtime power orchestration.

\textbf{Architecture and training task shape the observed PFI distribution.}
As shown in Fig.~\ref{fig:pfi}, we observe that dense jobs have the lowest PFI (median 1.27), MoE fine-tuning jobs have the highest PFI (median 2.10), and MoE pretraining jobs fall in between (median 1.43). Applying Dunn's test for statistical significance of group differences, we find support for MoE fine-tuning as a distinct cluster (corrected $p<0.05$) but are unable to distinguish MoE pretraining as a third cluster distinct from dense or MoE fine-tuning. Appendix~\ref{appendix:group_diff_tests} provides the results of our statistical group difference testing. 

\begin{figure}[tb]
  \centering
  \includegraphics[width=\textwidth]{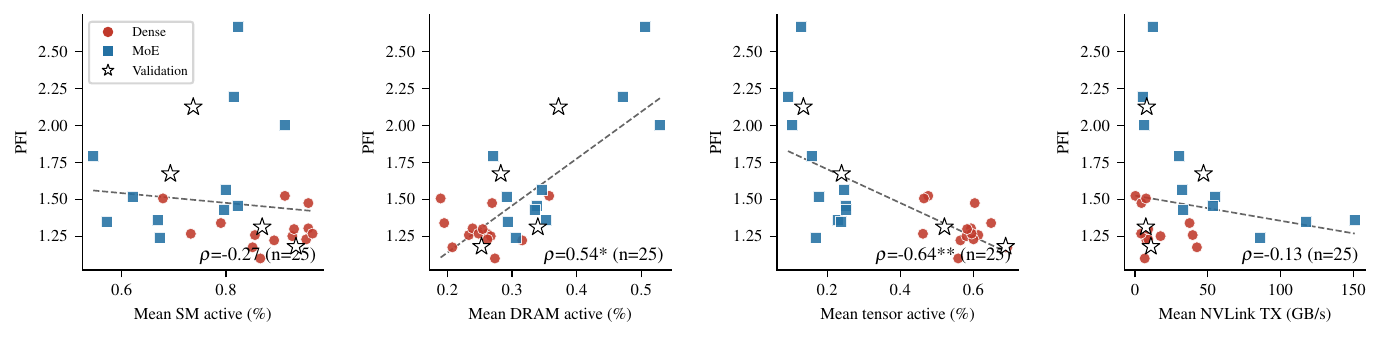}
  \includegraphics[width=\textwidth]{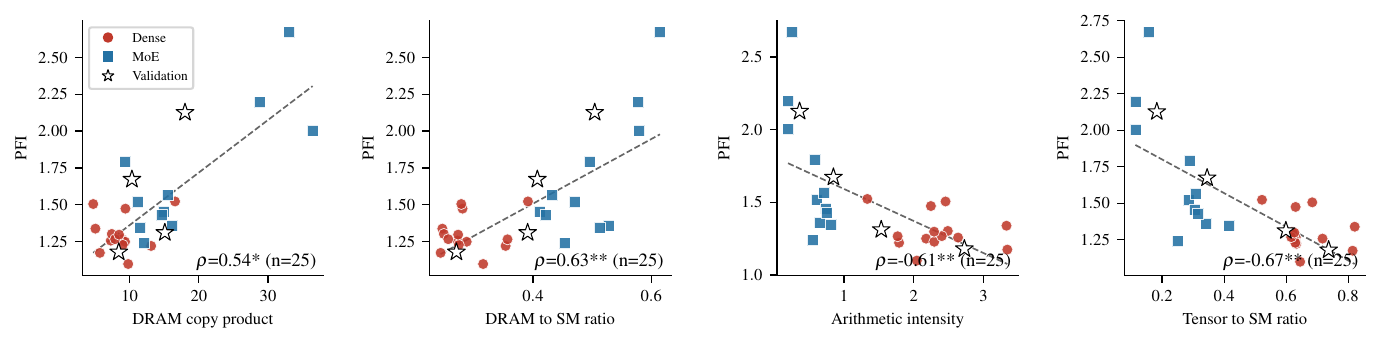}
  \caption{%
    \small{PFI vs.\ GPU telemetry metrics, showing the correlation (ordinary least-squares fit and Spearman $\rho$) between PFI and metrics measured for the uncapped job in each sweep group. DRAM copy product and tensor to SM ratio are derived metrics.
    $^{*}q<0.05$, $^{**}q<0.01$, where $q$ is the
    Benjamini--Hochberg FDR-corrected $p$-value across all metrics tested in Fig.~\ref{fig:appendix:all_metrics} in the appendix.}}
  \label{fig:pfi_vs_dcgm}
  \vspace{-0.2in}
\end{figure}

\textbf{Memory and compute intensity-related telemetry show the strongest association with PFI.}
Fig.~\ref{fig:pfi_vs_dcgm} shows the relationship between PFI and several metrics gathered through DCGM, including mean SM, DRAM, and tensor pipe activity, NVLink utilization, and selected composite metrics. We observe statistically significant positive correlation between PFI and memory-related signals (mean DRAM activity and DRAM copy product), as well as statistically significant negative correlation with tensor pipe activity.
We tested several additional composite metrics, reported in Appendix~\ref{appendix:more_plots}; however, due to the limited size of our dataset, we limit our analysis in \S\ref{sec:discussion} to features with a strong mechanistic link to power flexibility.

\section{Discussion}\label{sec:discussion}

\begin{figure}[b]
  \centering
  \begin{minipage}[t]{0.45\textwidth}
    \centering
    \includegraphics[width=\linewidth]{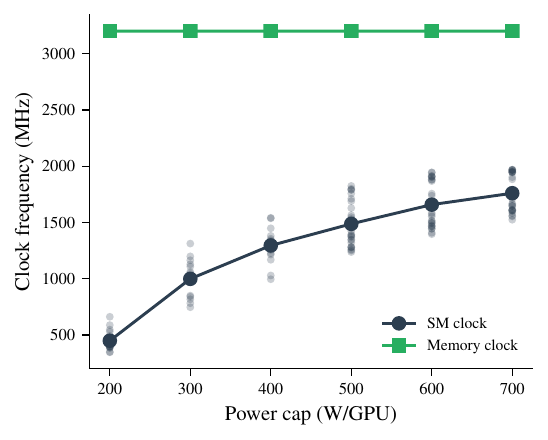}
  \caption{%
   \small{Steady-state SM clock and memory clock frequency
    as a function of power cap.}}
  \label{fig:clocks}
  \end{minipage}
  \hfill
  \begin{minipage}[t]{0.4\textwidth}
      \centering
      \includegraphics[width=\linewidth]{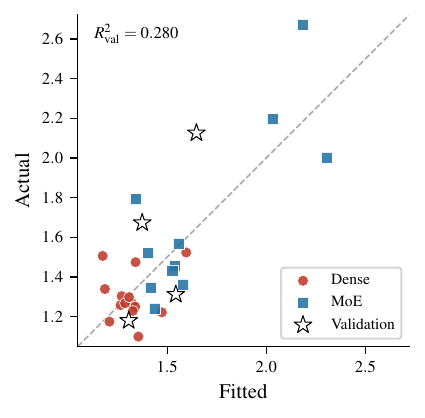}
        \vspace{-0.28in}
      \caption{%
        \small{Predictions of the PFI model fit to DRAM copy product, including the validation set.}}
    \label{fig:pfi_pred_actual}
  \end{minipage}
\end{figure}

To understand the mechanism behind the different PFI regimes in \S\ref{sec:results}, as well as the correlation between memory activity and PFI, we can examine how the H200 implements power caps. Fig.~\ref{fig:clocks} shows that the H200 caps power by preferentially reducing SM clock frequency while leaving memory clock frequency essentially unchanged: capped GPUs lose compute throughput, but retain memory bandwidth. Memory-bound jobs therefore absorb caps with a smaller throughput loss than compute-bound jobs. As shown in Fig.~\ref{fig:pfi_vs_dcgm}, MoE models exhibit higher values for memory-related telemetry, suggesting that these signals provide a measurable proxy for regime, separating compute-bound and memory-bound jobs.
While extensively studying whether this mechanism generalizes across accelerator architectures is beyond the scope of this work, Appendix~\ref{appendix:h100} includes a limited validation on H100s, which also preferentially throttle compute speed over memory bandwidth~\cite{ujeniya2026h100h200}, showing a similar pattern.

%

As defined in Eq.~\ref{eq:pfi}, computing PFI requires power-throughput measurements at a range of power caps, making it difficult to calculate for production jobs. While we find that PFI varies by job type, the power management layer typically does not know which architecture or task is being run on any given hardware node. As a result, the ability to correlate GPU telemetry with task/architecture regime (and thus PFI) is a critical capability for production use.

To build this estimator, we use the feature selection and ablation results presented in Table~\ref{tab:feature_ablation}. Specifically, we fit a least-squares linear model using two candidate features derived from DCGM data and motivated by the memory-bound mechanism of power flexibility discussed above: DRAM copy product (the product of DRAM activity and memory copy utilization) and arithmetic intensity (the ratio of tensor pipe activity to DRAM activity). We evaluate each model using both leave-one-out (LOO) and leave-one-cluster-out (LOCO, holding out one architecture-task cluster at a time), reporting in-sample and out-of-sample $R^2$, RMSE, and Spearman $\rho$. For completeness, we also show the results of fitting a model on all DCGM metrics as well as ground-truth job metadata (architecture, task, and GPU count).

We find that a simple linear model fit on DRAM copy product alone (shown in Fig.~\ref{fig:pfi_pred_actual}) provides the best performance across both LOO and LOCO, and we observe consistent $R^2$ values between the validation data and LOCO analysis. Consistent with our observation that PFI appears to cluster by job type (Dense vs. MoE, pretraining vs. fine-tuning), the ability of DCGM signals like DRAM copy product to predict PFI is likely because these signals provide both a measurable proxy for job type and a means of rank-ordering the expected relative PFI of different jobs. Appendix~\ref{app:dram_validation} provides further sensitivity analysis for this model.

Given the limited number of observations ($n=25$ samples of PFI derived from $131$ runs), we cannot justify higher-order models (e.g., nonlinear relationships or using more DCGM metrics), but future work may expand to larger sample sizes and different functional forms.

\begin{table}[tb]
  \centering
  \caption{%
    \small{Predictor ablation for PFI ($n=25$).
    Job metadata includes architecture type, training task, and $\log_2$ GPU count.
    $p$: \# of features.
    LOO: leave-one-out.
    LOCO: leave-one-cluster-out.
    $\rho$: Spearman rank correlation.}}
  \label{tab:feature_ablation}
  \small
\begin{tabular}{lrrrrrrrr}
    \toprule
    & & & \multicolumn{3}{c}{LOO} & \multicolumn{3}{c}{LOCO} \\
    \cmidrule(lr){4-6} \cmidrule(lr){7-9}
    Predictors & $p$ & $R^2_{\mathrm{in}}$ & $R^2$ & RMSE & $\rho$ & $R^2$ & RMSE & $\rho$ \\
    \midrule
    (i) DRAM copy product & 1 & 0.675 & \textbf{0.524} & \textbf{0.241} & \textbf{0.489} & \textbf{0.246} & \textbf{0.303} & \textbf{0.515} \\
    (ii) Arithmetic intensity & 1 & 0.381 & 0.238 & 0.305 & 0.482 & -1.652 & 0.569 & 0.311 \\
    (iii): (i) + (ii) & 2 & 0.676 & 0.462 & 0.256 & 0.435 & -0.588 & 0.440 & 0.357 \\
    (iv) Job metadata & 3 & 0.519 & 0.310 & 0.290 & 0.377 & -4.106 & 0.789 & 0.188 \\
    (v): (i) + (iv) & 4 & 0.687 & 0.350 & 0.282 & 0.386 & -2.391 & 0.643 & 0.208 \\
    (vi): All DCGM metrics & 13 & 0.885 & -0.187 & 0.381 & 0.323 & -18.483 & 1.542 & 0.077 \\
    (vii): All DCGM metrics + (iv) & 16 & 0.942 & -0.010 & 0.351 & 0.412 & -104 & 3.572 & -0.508 \\
    \bottomrule
  \end{tabular}
\end{table}

\section{Exploiting Power Flexibility Index}
\label{sec:exploit}

PFI is a control-oriented metric: our ultimate goal is to allow the infrastructure layer to maximize total tokens/sec throughput under a global power budget by protecting relatively inflexible jobs and assigning deeper power reductions to relatively flexible jobs. To demonstrate this capability, we construct simulated workloads by randomly sampling collections of jobs in our dataset and apply five different strategies to allocate power caps across these jobs:
\begin{itemize}
    \item \textit{Oracle}: optimal allocation using perfect knowledge of each job's throughput-power curve.
    \item \textit{Equal weight}: allocate power reductions proportionately to each job's uncapped power.
    \item \textit{MoE FT-weighted}: based on equal weight but doubles the power reduction allocated to MoE fine-tuning jobs (the highest flexibility cluster).
    \item \textit{PFI-aware}: adjusts the proportional allocation using PFI, as described below.
    \item \textit{Uncalibrated DCGM (DRAM-copy-product)}: allocate power reductions proportionately to each job's DRAM copy product, showing the contrast between the calibrated PFI model and raw telemetry.
\end{itemize}

Denote the power reduction applied to job $i$ as $P_{\rm{curtail}}^i$, the uncapped power and throughput of job $i$ as $P^i_{\rm{max}}$ and $T^i_{\rm{max}}$, and the PFI of job $i$ as $PFI^i$,
then:
\begin{equation}
  \text{Equal weight: } P_{\rm curtail}^i \propto P^i_{\rm max}
  \qquad\qquad
  \text{PFI-aware: } P_{\rm curtail}^i \propto \frac{P^i_{\rm max}}{T^i_{\rm max}} PFI^i
\end{equation}

We simulate the performance of these strategies on 500 synthetic workloads in two scenarios, each with 100 jobs. In the ``random'' scenario, we sample jobs with replacement from our dataset. In the ``production'' scenario, we sample jobs according to the pretraining/fine-tuning ratio provided in recent work~\cite{huCharacterization2024}. We estimate PFI using the model from \S\ref{sec:discussion}, 
fitted to telemetry measured on the uncapped run of each job, mirroring the data that would be available from uncapped jobs running in production. We then allocate cluster-level GPU power reductions between 0\% and 40\% across jobs using each strategy and record the corresponding throughput reduction (simulated by interpolating the measured power-throughput curve for each job).

\begin{figure}[t]
    \centering
    \includegraphics[width=\linewidth]{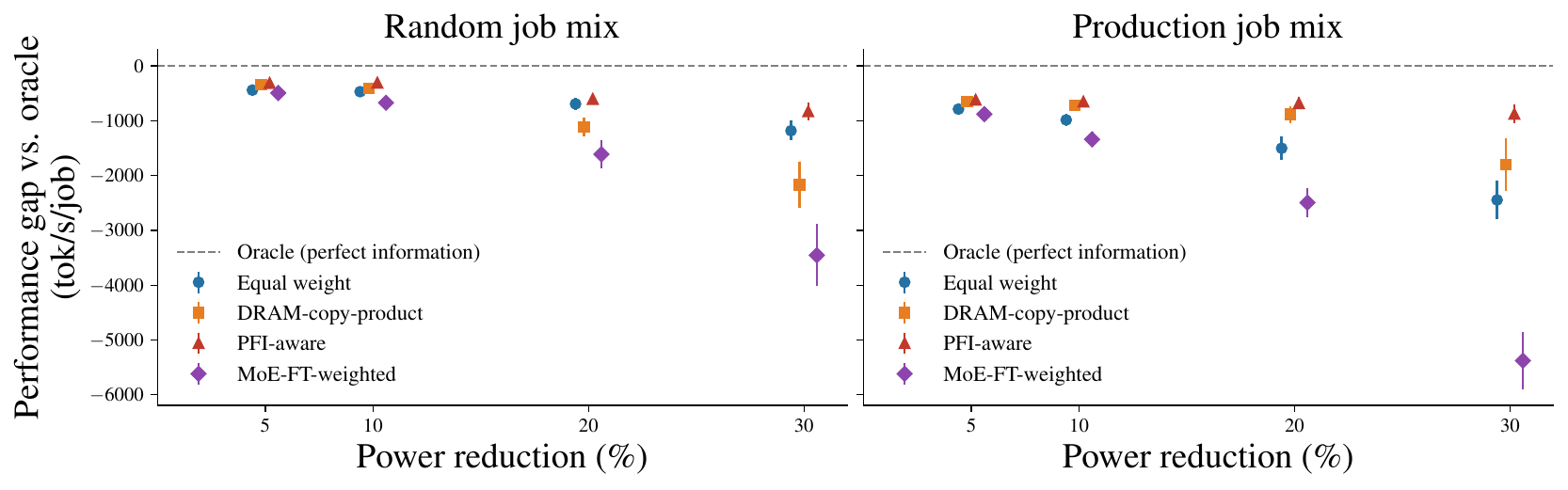}
  \caption{%
    \small{Difference in throughput for four allocation strategies compared against an oracle with perfect information of each job's power-performance curve (less negative = closer to optimal) across a range of curtailment targets and workload mixes (random = jobs sampled at random from our dataset, production = jobs sampled according to the pretraining/fine-tuning mix from recent work~\cite{huCharacterization2024}).}}
  \label{fig:orchestration}
\end{figure}

Fig.~\ref{fig:orchestration} shows that the PFI-aware strategy yields equal or higher throughput than the equal weight strategy across all power reduction levels and both workload mix scenarios. While the random job mix includes a roughly equal mix of pretraining and fine-tuning, the production mix~\cite{huCharacterization2024} includes a large number of small fine-tuning jobs (which tend to have high PFI) and a small number of large pretraining jobs (which tend to be lower PFI).
The PFI-aware orchestration strategy is better able to exploit the differences in power elasticity between these job types, leading to substantially improved performance with the production job mix. Under a 30\% power reduction on the production mix, PFI-aware power allocation recovers $\sim$1.5k tokens/s per job, 63\% of the performance gap between an equal-weight allocation and an oracle with perfect information.
These results show how PFI can enable power-aware orchestration of LLM training workloads, providing the foundation for data center operators to offer flexible SLAs while preserving token throughput.

\section{Related Work}
\input{relatedwork}

\section{Limitations}\label{sec:limitations}

Our characterization is scoped to NVIDIA H200 GPUs on \texttt{a3-ultragpu-8g} nodes, uses \texttt{nvidia-smi -pl} as the sole power-control mechanism, and covers pretraining and rank-32 LoRA SFT runs of 30 min each.
The 32-GPU runs include only \texttt{Qwen3-32B} pretraining, since rank-32 LoRA SFT is rarely deployed at this scale and our 32-GPU capacity was constrained; broader scaling across architectures remains a natural extension.
PFI is computed from aggregate GPU power only: host CPU/DRAM, NIC, PSU, fan, and rack-level cooling are not measured, and broadening the power signal in Eq.~\ref{eq:pfi} as those streams become available is a direct extension.

Each PFI data point represents a single power sweep; a subset of 14 configurations was repeated (35 runs total) and within those repeats steady-state aggregate GPU power and tokens/s agreed with median coefficient of variation 0.4\% and 0.5\%.
The $n=25$ PFI sample size still limits cross-architecture generalization, which we mitigate via held-out validation models.
The curtailment simulation in §\ref{sec:exploit} uses the same power-throughput curves used to measure PFI; validation on out-of-sample workloads is left to future work with a larger dataset.
Closed-loop scheduler validation of §\ref{sec:exploit} and extensions to other accelerators (including heterogeneous clusters), RLHF, and inference are left to future work.

\section{Conclusion}

While AI data centers can operate as power-aware, grid-responsive loads, doing so while maximizing training throughput requires a principled understanding of how LLM training workloads respond to sustained reductions in available GPU power. In this work, we show that training-job power elasticity can be quantified, estimated at runtime, and exploited for intelligent power-aware workload orchestration.

\paragraph{Broader impact}
Rapid growth in LLM training is placing significant new demands on electric power systems and is increasingly constrained by limited near-term power availability. By enabling data center operators to reduce power consumption when needed while maximizing throughput and minimizing disruption to customer workloads, this work supports a new class of power-flexible AI infrastructure that can improve utilization of constrained grid capacity, support faster AI deployment timelines, and reduce the infrastructure cost of large-scale AI expansion.




\begin{ack}
We would like to acknowledge Fatih Acun and Brian Kulis at Emerald AI for their support and constructive feedback during the review phase.
\end{ack}


\newpage
\bibliographystyle{abbrv}
\bibliography{references}
\medskip

\clearpage
\appendix

\section{Alternative definitions of the power flexibility index (PFI)}\label{appendix:alternative_pfi}

We consider three candidate definitions of PFI and discuss their tradeoffs.

\paragraph{Ratio (adopted definition)}
Define PFI at each power cap,
\begin{equation}
    PFI_i = \frac{1 - P_i/P_{\max}}{1 - T_i/T_{\max}},
\end{equation}
and take the average over measured caps: $PFI = \frac{1}{N}\sum_{i=1}^N PFI_i$ (excluding the uncapped point).
This definition is simple to compute and has a direct interpretation as the ratio of
fractional power reduction to fractional throughput reduction. Its main limitation is
sensitivity to the choice of sweep points: if different jobs are measured at
different power caps, the averages are not directly comparable. However, Appendix~\ref{app:grid} shows that changing the sweep grid does not change the PFI results significantly (the relative ordering of PFI is preserved under different sweep grids). 

\paragraph{Area between curves}
Define PFI as the area between the normalized power--throughput curve and the unit-slope
diagonal,
\begin{equation}
    PFI = \int_0^1 \left(\frac{T}{T_{\max}} - \frac{P}{P_{\max}}\right)
          \frac{dP}{P_{\max}},
\end{equation}
approximated with the trapezoid rule:
\begin{equation}
    PFI \approx \frac{1}{2 T_{\max} P_{\max}}
    \sum_{i=1}^{N} \left[(T_i + T_{i-1})(P_i - P_{i-1})\right] - \frac{1}{2},
\end{equation}
with boundary condition $(P_0, T_0) = (0, 0)$. This definition is robust to the choice
of sample points, but it is less intuitive than the ratio
definition.

\begin{figure}[b]
  \centering
  \includegraphics[width=0.4\textwidth]{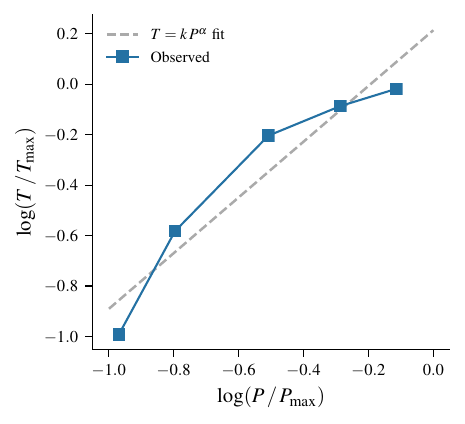}
  \caption{%
   \small{ Comparison of constant-elasticity fit with the observed power-performance curve for \texttt{gpt-oss-20b} fine-tuning on 8$\times$H200s.}}
  \label{fig:constant_elasticity_fit}
\end{figure}

\paragraph{Constant-elasticity model}
Fit the log-linear model $\log T = \alpha \log P + \log k$ and define $PFI = 1 -
\alpha$, so that the implied power--throughput relationship is $T = k P^{1-PFI}$.
Under this parameterization, $PFI = 0$ corresponds to throughput proportional to power
(fully inflexible), and $PFI = 1$ corresponds to throughput independent of power (fully
flexible). The model has a single degree of freedom per job, making it well-suited
to sweeps with few power caps; however, we found that this constant-elasticity model was a poor fit to our observed data, as shown in Fig.~\ref{fig:constant_elasticity_fit}.

\begin{figure}[tb]
  \centering
  \includegraphics[width=0.7\textwidth]{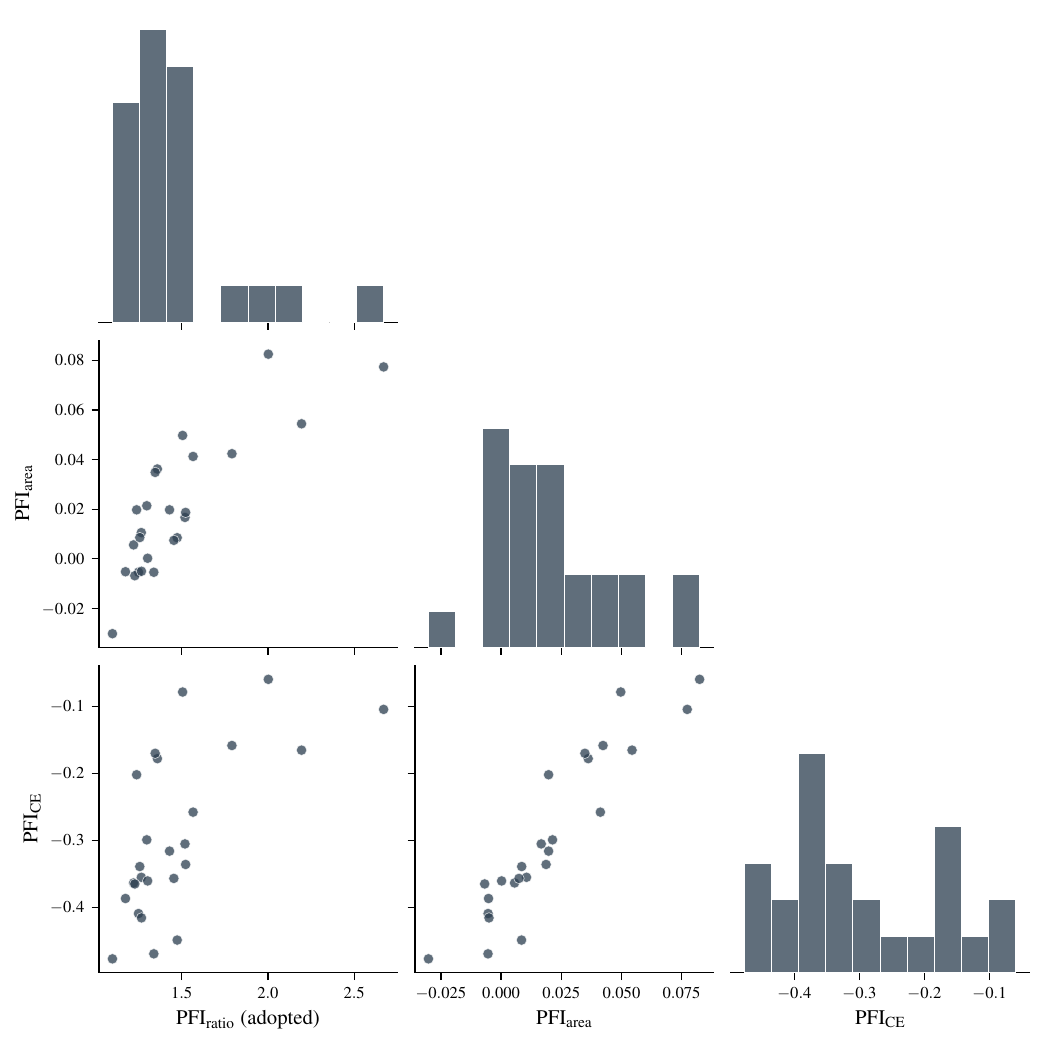}
  \caption{%
\small{Comparison of the empirical distribution of the three alternative PFI definitions on the training set cohort.}}
  \label{fig:pfi_definition_pairplot}
\end{figure}

Empirically, we find that all three options are highly correlated (see Fig.~\ref{fig:pfi_definition_pairplot}), but we select the proposed ratio definition for two reasons. First, it has a direct interpretation as ratio of power reduction to throughput loss, and the physical interpretation of the other two metrics is less clear.
Second, it provides a clear segmentation between power inelastic (PFI $> 1$) and elastic (PFI $\leq 1$), with a dynamic range ($\approx [1, 2.5]$). The area definition also segments (at zero), but we found that it has a very narrow range ($\approx[-0.1, 0.1]$).

\section{Detailed experimental setup}\label{appendix:detailed-setup}

\subsection{Infrastructure}

We ran all experiments on Google Kubernetes Engine (GKE) using \texttt{a3-ultragpu-8g} nodes with 8$\times$ NVIDIA H200 GPUs (141\,GiB HBM3e, 700\,W TDP), fourth-generation NVLink, and a CX-7 RoCE fabric exposing eight GPUDirect-RDMA NICs per node over an MTU-8896 multi-VPC topology.
Multi-node runs (16- and 32-GPU) are scheduled through Kueue with Dynamic Workload Scheduler (DWS) Flex-start or Spot capacity.
We provisioned clusters in multiple GCP regions based on availability of Flex-start and Spot allocations.

\subsection{Software stack}

Workers ran a Docker image (\texttt{rayproject/ray:2.53.0-py312-cu128}) with Ray~2.53.0, PyTorch~2.10 (CUDA~12.8), Transformers~$\geq 5.2$, Accelerate~$\geq 1.10$, PEFT~$\geq 0.17$, FlashAttention~2.8.1 (\texttt{cu12/torch2.10} prebuilt wheel), \texttt{bitsandbytes}~$\geq 0.43.3$, and \texttt{torchtitan}~0.2.2.
Distributed training uses Ray Train's \texttt{TorchTrainer} with NCCL backend; FSDP (full-shard, auto-wrap by transformer-decoder layer class) is used for dense jobs and for all LoRA SFT jobs.
MoE pretraining is dispatched to a torchtitan subprocess that drives FSDP2 with native Expert Parallelism.
We use a flat parallelism mesh: \texttt{data\_parallel\_shard\_degree} equals \texttt{world\_size} (FSDP shards across every rank), with expert parallelism nested inside the FSDP group at degree \texttt{EP}.

GPU telemetry is harvested by an \texttt{nvcr.io/nvidia/k8s/dcgm-exporter:3.3.7-3.5.0-ubuntu\\22.04} sidecar deployed alongside every worker pod and accessed from the Ray container. Prior to collecting data, we verified that logging telemetry using DCGM does not materially affect job performance.

\subsection{Models and datasets}


Pretraining and SFT runs use FineWeb-Edu
and UltraChat-200K,
respectively, pre-tokenized into Ray Data shards in region-local object storage. 
\texttt{gpt-oss-20b} is dequantized from MXFP4-packed expert weights to BF16.
We use the AdamW optimizer with default parameters.


\subsection{Power control mechanism}


Per-GPU power caps are set using \texttt{nvidia-smi -i <local\_rank> -pl <watts>}, called at the start of every run (after \texttt{torch.cuda.set\_device} but before any CUDA context allocation).
The cap is swept in 100 \,W increments from 200\,W to 700\,W on single-node (8-GPU) groups; the 16- and 32-GPU groups sweep 200\,W, 500\,W, 600\,W, and 700\,W only.
We verify that each power cap was set correctly by reading \texttt{DCGM\_FI\_DEV\_POWER\_MGMT\_LIMIT} and \texttt{DCGM\_FI\_DEV\_ENFORCED\_POWER\_LIMIT} from DCGM during the steady-state phase of each run.

\subsection{Performance data collection}


GPU metrics are scraped from the DCGM sidecar at 5\,s intervals by a single \texttt{local\_rank=0} worker per node. System metrics (CPU, RSS, disk and network bytes) are sampled in-process via \texttt{psutil}, also every 5\,s.
Training metrics are logged on every step by a custom \texttt{TelemetryCallback} and include global step, loss, learning rate, observed tokens-per-second, achieved TFLOPS, and Model FLOPs Utilization (MFU).
RDMA counters from the kernel's InfiniBand interface are appended on RoCE-equipped nodes when present.

\subsection{Measurement protocol}

Each configuration is run for 30\,minutes.
On the HuggingFace Trainer path (dense and LoRA jobs), evaluation is triggered at 10\,minutes and a single distributed-checkpoint write at 20\,minutes via an all-reduce-synchronized timer callback.
The torchtitan EP path triggers eval and checkpoint after fixed step counts chosen so that both events occur at similar times to the dense/LoRA jobs (10 and 20 minutes, respectively).
To isolate steady state from transients, we partition traces into three phases: \emph{warmup} (frame-buffer usage below 10\,GiB), \emph{startup} (weights loaded but \texttt{gpu\_util}~$=$~0), and \emph{training} (sustained \texttt{gpu\_util}~$\geq$~50\%).
All aggregates are reported over the training phase only.
Per-step throughput, MFU, and per-GPU power are aggregated as the mean of the top 10\% of samples within the training phase to exclude checkpoints and evals.
Runs at different power caps or hyperparameter values are executed sequentially on a single cluster after a 60 second delay, after verifying that GPU memory has been released.

\subsection{Compute resources}\label{sec:compute_resources}


The $131$ headline runs and $24$ held-out validation runs that produce the figures and tables in this paper consume approximately $1{,}000$ H200 GPU-hours, broken down by cluster size in Table~\ref{tab:compute_reported}.

\begin{table}[h]
\centering
\caption{\small{Compute for paper-included runs.
Wall-clock figures are the actual DCGM-bracketed durations from the master telemetry frame, not the $30$ min training cap.}}
\label{tab:compute_reported}
\begin{tabular}{lrrr}
\toprule
GPUs / cluster & Runs & Mean wall-clock (min) & GPU-hours \\
\midrule
8  & 120 & 34.9 & 558 \\
16 & 19  & 36.8 & 186 \\
32 & 16  & 38.1 & 325 \\
\midrule
Total & 155 & 35.6 & $\sim$1,069 \\
\bottomrule
\end{tabular}
\end{table}

The full project consumed additional H200 compute due to preliminary sweeps, hyperparameter tuning, and reruns of jobs that failed mid-training.
Across all six profiling regions, $410$ distinct H200 submissions produced DCGM telemetry and together consumed approximately $1{,}725$ GPU-hours (Table~\ref{tab:compute_total}). The $155$ reported runs are roughly one third of those submissions and account for $62 \%$ of H200 GPU-hours.

\begin{table}[h]
\centering

\caption{\small{Compute resources used for this project.}}
\label{tab:compute_total}
\begin{tabular}{lrr}
\toprule
Cluster shape & DCGM-active runs & GPU-hours \\
\midrule
H200, 8 GPUs   & 324 & $\sim$969 \\
H200, 16 GPUs  & 53  & $\sim$272 \\
H200, 32 GPUs  & 33  & $\sim$485 \\
\midrule
Total H200 (DCGM-active) & 410 & $\sim$1{,}725 \\
Reported in paper        & 155 & $\sim$1{,}069 \\
\midrule
H100, 8 GPUs (reported in Appendix~\ref{appendix:h100}) & 34 & $\sim$149 \\
\bottomrule
\end{tabular}

\end{table}
\section{Data tables}\label{appendix:data_tables}

Tables~\ref{tab:pfi_all_groups} and~\ref{tab:pfi_validation_groups} provide the raw PFI data for each sweep group in the headline and validation sets, respectively, including a 95\% confidence interval derived using Monte Carlo resampling from the empirical noise distributions derived from intra-run time-series variation (median standard error of the mean 0.04\% for power and 0.05\% for throughput).

\begin{figure}
  \centering
  \includegraphics[width=0.9\textwidth]{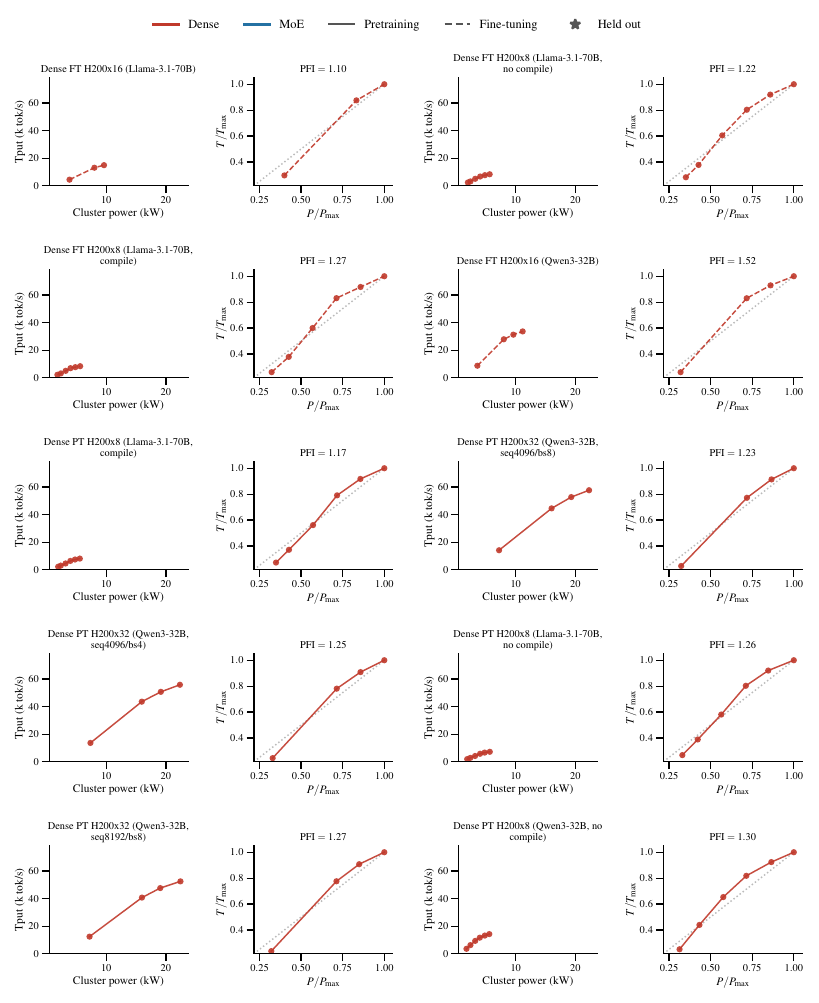}
  \caption{%
 \small{Power-throughput curves (raw and normalized) for each headline and validation run on H200.}}
  \label{fig:all_pfi_curves_1}
\end{figure}

\begin{figure}
  \centering
  \includegraphics[width=0.9\textwidth]{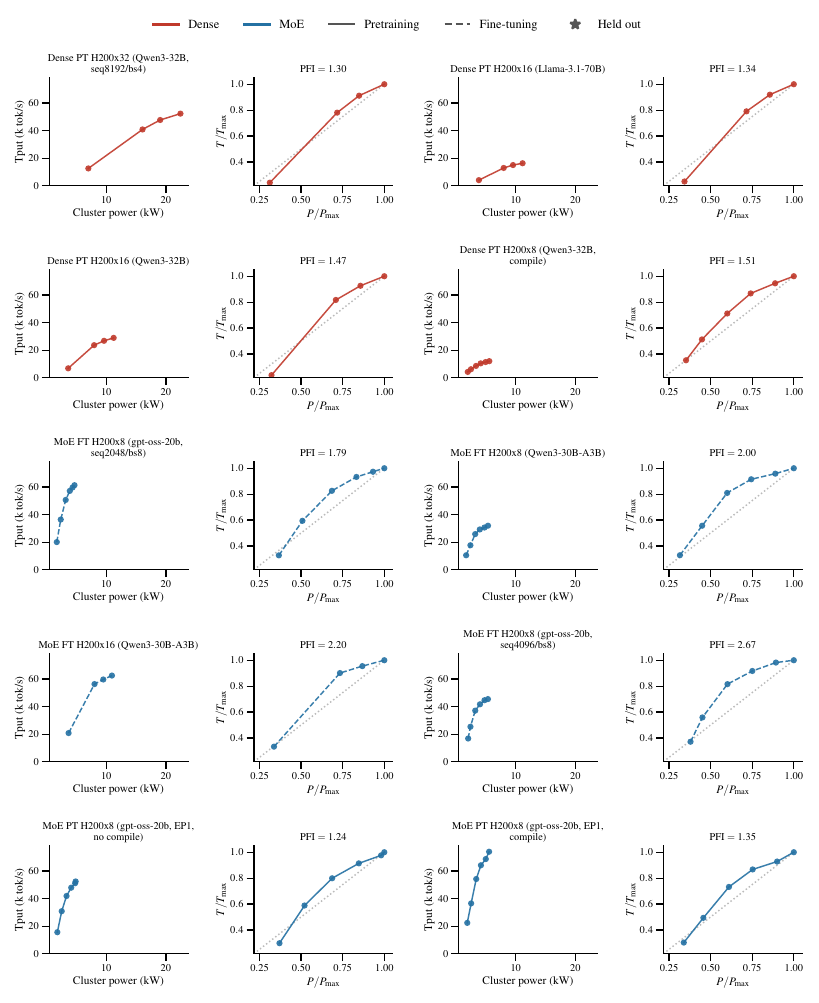}
  \caption{%
 \small{Power-throughput curves (raw and normalized) for each headline and validation run on H200 (continued).}}
  \label{fig:all_pfi_curves_2}
\end{figure}

\begin{figure}
  \centering
  \includegraphics[width=0.9\textwidth]{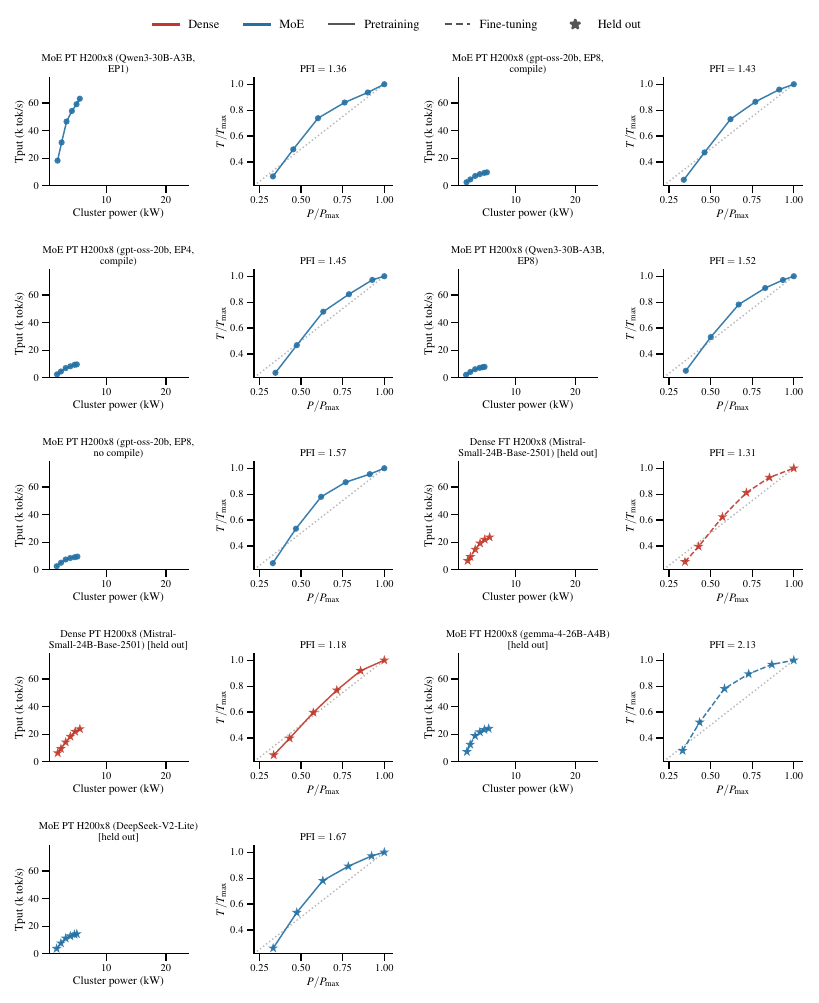}
  \caption{%
 \small{Power-throughput curves (raw and normalized) for each headline and validation run on H200 (continued).}}
  \label{fig:all_pfi_curves_3}
\end{figure}

In addition, Figs.~\ref{fig:all_pfi_curves_1}, \ref{fig:all_pfi_curves_2}, and~\ref{fig:all_pfi_curves_3} includes the raw and normalized power-performance curves for each job in Tables~\ref{tab:pfi_all_groups} and~\ref{tab:pfi_validation_groups}.

\begin{table}[t]
  \centering
  \caption{Power Flexibility Index for each of the headline sweep groups on H200. Each group runs a single job (fixed model, task, cluster size, and training configuration) at multiple power caps; \# runs is the number of power-caps measured. 95\% CI is estimated from 10,000 Monte Carlo samples for each run, with power and throughput measurements resampled from the empirical noise distribution.}
  \label{tab:pfi_all_groups}
  \begin{tabular}{lrrc}
    \toprule
    Job & \# runs & PFI & 95\% CI \\
    \midrule
    Dense FT H200x16 (Llama-3.1-70B) & 3 & 1.10 & $[1.096,\,1.102]$ \\
    Dense FT H200x8 (Llama-3.1-70B, no compile) & 6 & 1.22 & $[1.218,\,1.225]$ \\
    Dense FT H200x8 (Llama-3.1-70B, compile) & 6 & 1.27 & $[1.264,\,1.269]$ \\
    Dense FT H200x16 (Qwen3-32B) & 4 & 1.52 & $[1.519,\,1.527]$ \\
    Dense PT H200x8 (Llama-3.1-70B, compile) & 6 & 1.17 & $[1.171,\,1.177]$ \\
    Dense PT H200x32 (Qwen3-32B, seq4096/bs8) & 4 & 1.23 & $[1.226,\,1.230]$ \\
    Dense PT H200x32 (Qwen3-32B, seq4096/bs4) & 4 & 1.25 & $[1.248,\,1.251]$ \\
    Dense PT H200x8 (Llama-3.1-70B, no compile) & 6 & 1.26 & $[1.254,\,1.260]$ \\
    Dense PT H200x32 (Qwen3-32B, seq8192/bs8) & 4 & 1.27 & $[1.265,\,1.270]$ \\
    Dense PT H200x8 (Qwen3-32B, no compile) & 6 & 1.30 & $[1.295,\,1.301]$ \\
    Dense PT H200x32 (Qwen3-32B, seq8192/bs4) & 4 & 1.30 & $[1.300,\,1.305]$ \\
    Dense PT H200x16 (Llama-3.1-70B) & 4 & 1.34 & $[1.335,\,1.342]$ \\
    Dense PT H200x16 (Qwen3-32B) & 4 & 1.47 & $[1.470,\,1.478]$ \\
    Dense PT H200x8 (Qwen3-32B, compile) & 6 & 1.51 & $[1.494,\,1.518]$ \\
    MoE FT H200x8 (gpt-oss-20b, seq2048/bs8) & 6 & 1.79 & $[1.750,\,1.833]$ \\
    MoE FT H200x8 (Qwen3-30B-A3B) & 6 & 2.00 & $[1.976,\,2.028]$ \\
    MoE FT H200x16 (Qwen3-30B-A3B) & 4 & 2.20 & $[2.146,\,2.244]$ \\
    MoE FT H200x8 (gpt-oss-20b, seq4096/bs8) & 6 & 2.67 & $[2.615,\,2.729]$ \\
    MoE PT H200x8 (gpt-oss-20b, EP1, no compile) & 6 & 1.24 & $[1.209,\,1.268]$ \\
    MoE PT H200x8 (gpt-oss-20b, EP1, compile) & 6 & 1.35 & $[1.335,\,1.358]$ \\
    MoE PT H200x8 (Qwen3-30B-A3B, EP1) & 6 & 1.36 & $[1.349,\,1.370]$ \\
    MoE PT H200x8 (gpt-oss-20b, EP8, compile) & 6 & 1.43 & $[1.407,\,1.454]$ \\
    MoE PT H200x8 (gpt-oss-20b, EP4, compile) & 6 & 1.45 & $[1.435,\,1.476]$ \\
    MoE PT H200x8 (Qwen3-30B-A3B, EP8) & 6 & 1.52 & $[1.456,\,1.597]$ \\
    MoE PT H200x8 (gpt-oss-20b, EP8, no compile) & 6 & 1.57 & $[1.557,\,1.576]$ \\
    \bottomrule
  \end{tabular}
\end{table}

\begin{table}[t]
  \centering
  \caption{Power Flexibility Index for the 4 held-out validation sweep groups on H200.}
  \label{tab:pfi_validation_groups}
  \begin{tabular}{lrr}
    \toprule
    Job & \# runs & PFI \\
    \midrule
    Dense FT H200x8 (Mistral-Small-24B-Base-2501) & 6 & 1.31 \\
    Dense PT H200x8 (Mistral-Small-24B-Base-2501) & 6 & 1.18 \\
    MoE FT H200x8 (gemma-4-26B-A4B) & 6 & 2.13 \\
    MoE PT H200x8 (DeepSeek-V2-Lite) & 6 & 1.67 \\
    \bottomrule
  \end{tabular}
\end{table}


\section{Additional results}

\subsection{All DCGM metrics}\label{appendix:more_plots}

\begin{figure}[h]
  \centering
  \includegraphics[width=0.9\textwidth]{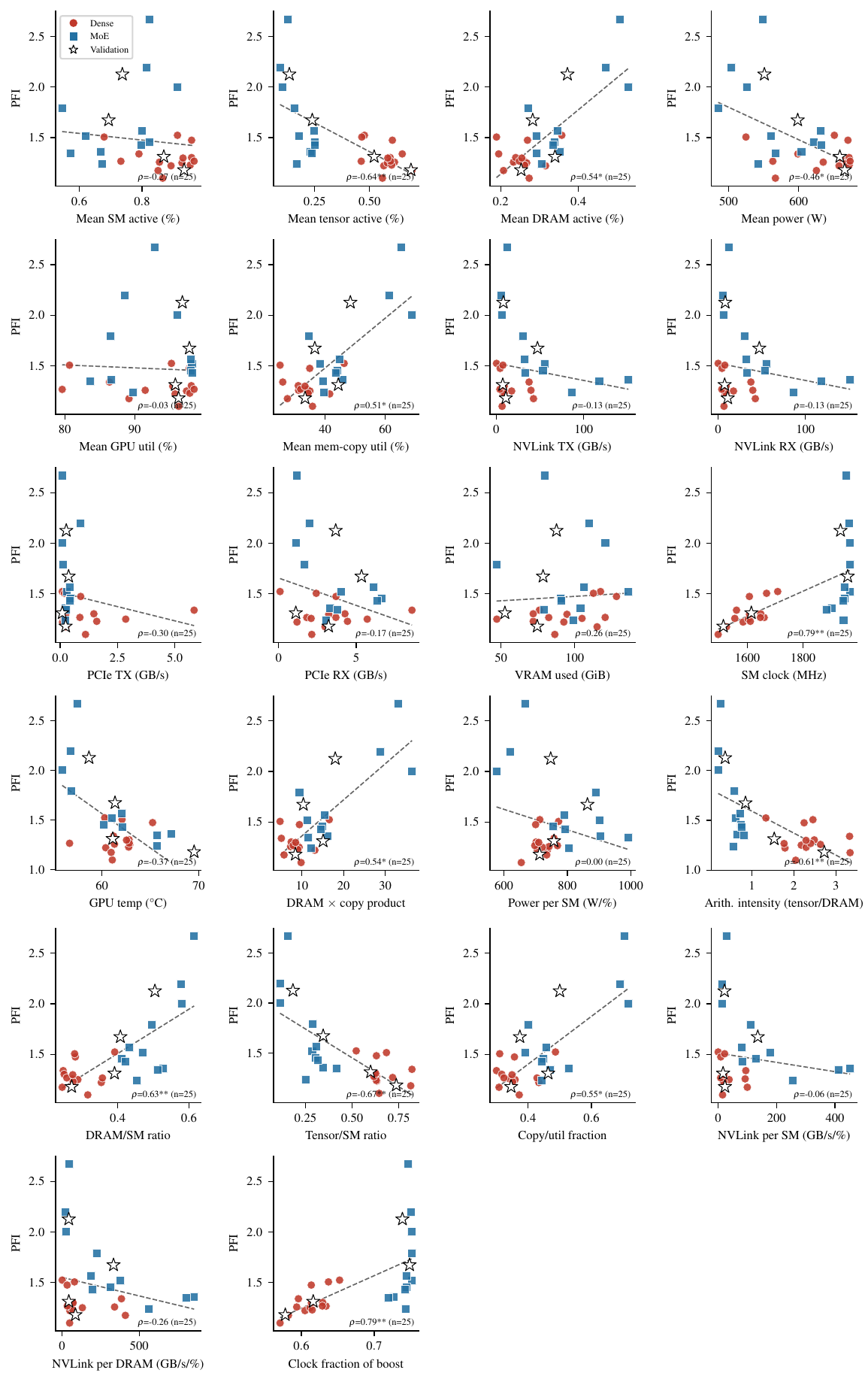}
  \caption{%
 \small{   PFI vs. all DCGM GPU metrics and composite features. Dashed lines show ordinary least-squares fit; Spearman $\rho$ and significance are annotated. $^{*}q<0.05$, $^{**}q<0.01$, where $q$ is the
    Benjamini--Hochberg FDR-corrected $p$-value across all metrics in this figure.}}
  \label{fig:appendix:all_metrics}
\end{figure}

For completeness, Fig.~\ref{fig:appendix:all_metrics} extends Figs.~\ref{fig:pfi_vs_dcgm} to all measured DCGM metrics.

\subsection{H100 validation}\label{appendix:h100}

Our main characterization is confined to H200 GPUs. To test whether the
qualitative patterns we report depend on that specific platform, we ran a
small set of matched sweeps on H100 GPUs: six sweep groups (34 runs total),
with at least one group in each of the four architecture$\times$task cells
(dense PT, dense FT, MoE PT, MoE FT). Table~\ref{tab:h100} reports these results along with those of the matched H200 sweeps.

The rank ordering of PFI is largely preserved between H100 and H200
(Spearman $\rho = 0.943$), and we observe that the DCGM-to-PFI model fit on H200 data achieves $R^2=0.56$ on the H100 data. The largest
absolute difference in PFI between H100 and H200 is $0.16$ and the median difference is $3.5\%$, which is small compared to
the between-job PFI range of $1.10$ to $2.67$ on H200. We observe no
systematic direction to the difference in PFI between platforms. The sample size presented here is not sufficient to claim generalizable results on H100 workloads.


\begin{table}[t]
  \centering
  \caption{PFI for six matched sweep groups on H100 and H200. All groups use
  8 GPUs. Rank ordering is nearly preserved across generations (Spearman
  $\rho = 0.943$).}
  \label{tab:h100}
  \small
  \begin{tabular}{@{}llcccc@{}}
    \toprule
    Arch/Task & Model & EP & Runs (H100) & PFI (H200) & PFI (H100) \\
    \midrule
    Dense PT & Mistral-Small-24B-Base-2501 & 1 & 6           & 1.18 & 1.26 \\
    Dense FT & Llama-3.1-70B               & 1 & 4$^{\ast}$  & 1.22 & 1.19 \\
    Dense FT & Mistral-Small-24B-Base-2501 & 1 & 6           & 1.31 & 1.33 \\
    MoE PT   & gpt-oss-20b                 & 8 & 6           & 1.43 & 1.38 \\
    MoE PT   & DeepSeek-V2-Lite            & 8 & 6           & 1.67 & 1.51 \\
    MoE FT   & Qwen3-30B-A3B               & 1 & 6           & 2.00 & 1.93 \\
    \bottomrule
  \end{tabular}

  \vspace{0.5em}
  \begin{minipage}{0.95\linewidth}
    \footnotesize
    $^{\ast}$This group contains four runs rather than six: the 500\,W and
    200\,W runs failed with transient errors and could not be re-run within
    the available allocation.
  \end{minipage}
\end{table}

\subsection{Sensitivity of PFI to the Power-Cap Sweep Grid}
\label{app:grid}

PFI averages the ratio of fractional power reduction to fractional
throughput reduction over the sampled power caps, so its value depends on
which caps are sampled. Because our 8-GPU groups sweep six caps
(200--700\,W) while the 16- and 32-GPU groups sweep only four (200, 500,
600, 700\,W), comparing PFI between these groups confounds scale with the sweep grid. Table~\ref{tab:grid} quantifies the size of this effect by providing
PFI for every 8-GPU group recomputed using only the four caps present in the coarse
grid used for 16- and 32-GPU groups (excluding the 300\,W and 400\,W measurements). Restricting to the coarse grid increases PFI for all but one group (median increase $0.154$, or $10.7\%$) but does not materially change the ordering of groups (Spearman $\rho = 0.940$).

We draw two conclusions. First, absolute PFI values are comparable only
between groups measured on the same grid, and we accordingly avoid
quantitative comparisons of PFI magnitude between the 8-GPU groups and the
16-/32-GPU groups. Second, the rank ordering that our allocation policy
relies on is robust to this choice of grid, so PFI remains usable as an
ordinal signal across groups measured at different resolutions. While the PFI predictor presented in \S\ref{sec:discussion} is trained on the full dataset, our end-to-end results on power allocation in \S\ref{sec:exploit} demonstrate that this predictor is still useful in practice (likely because PFI remains correlated across different grids).

\begin{table}[t]
  \centering
  \caption{PFI for 8-GPU sweep groups computed on the full six-cap grid
  (200--700\,W) and on the coarse four-cap grid (200, 500, 600, 700\,W) used
  for the 16- and 32-GPU groups.}
  \label{tab:grid}
  \small
\begin{tabular}{lrrrr}
    \toprule
    Sweep group & PFI (6 caps) & PFI (4 caps) & $\Delta$ & $\Delta$ (\%) \\
    \midrule
    Dense FT H200x8 (Llama-3.1-70B, no compile) & 1.22 & 1.37 & $+0.145$ & $+11.9$ \\
    Dense FT H200x8 (Llama-3.1-70B, compile) & 1.27 & 1.44 & $+0.177$ & $+14.0$ \\
    \addlinespace
    Dense PT H200x8 (Llama-3.1-70B, compile) & 1.17 & 1.33 & $+0.155$ & $+13.2$ \\
    Dense PT H200x8 (Llama-3.1-70B, no compile) & 1.26 & 1.43 & $+0.177$ & $+14.1$ \\
    Dense PT H200x8 (Qwen3-32B, no compile) & 1.30 & 1.42 & $+0.121$ & $+9.3$ \\
    Dense PT H200x8 (Qwen3-32B, compile) & 1.51 & 1.67 & $+0.163$ & $+10.8$ \\
    \addlinespace
    MoE FT H200x8 (gpt-oss-20b, seq2048/bs8) & 1.79 & 1.98 & $+0.189$ & $+10.5$ \\
    MoE FT H200x8 (Qwen3-30B-A3B) & 2.00 & 2.22 & $+0.221$ & $+11.1$ \\
    MoE FT H200x8 (gpt-oss-20b, seq4096/bs8) & 2.67 & 3.32 & $+0.646$ & $+24.2$ \\
    \addlinespace
    MoE PT H200x8 (gpt-oss-20b, EP1, no compile) & 1.24 & 1.16 & $-0.082$ & $-6.6$ \\
    MoE PT H200x8 (gpt-oss-20b, EP1, compile) & 1.35 & 1.40 & $+0.055$ & $+4.1$ \\
    MoE PT H200x8 (Qwen3-30B-A3B, EP1) & 1.36 & 1.40 & $+0.037$ & $+2.7$ \\
    MoE PT H200x8 (gpt-oss-20b, EP8, compile) & 1.43 & 1.57 & $+0.144$ & $+10.1$ \\
    MoE PT H200x8 (gpt-oss-20b, EP4, compile) & 1.45 & 1.65 & $+0.194$ & $+13.3$ \\
    MoE PT H200x8 (Qwen3-30B-A3B, EP8) & 1.52 & 1.67 & $+0.153$ & $+10.1$ \\
    MoE PT H200x8 (gpt-oss-20b, EP8, no compile) & 1.57 & 1.66 & $+0.092$ & $+5.8$ \\
    \bottomrule
  \end{tabular}
\end{table}

\subsection{Statistical tests for group difference}\label{appendix:group_diff_tests}

We test whether PFI differs across task/architecture regimes using the $n = 25$ headline sweep
groups. Because dense pretraining and dense LoRA SFT are not separable in our data, we
pool them and compare three groups: dense, MoE pretraining, and MoE fine-tuning. We use
the Kruskal--Wallis $H$ test followed by
Dunn's post-hoc test over all three pairs with Holm--Bonferroni correction ($m = 3$).
The Kruskal--Wallis test rejects equality of the three groups ($H = 12.37$, $\mathrm{df} = 2$,
$p = 0.0021$). For Dunn's test, only the dense vs. MoE fine-tuning difference survives $p$-value correction. MoE pretraining is not separable from either dense or MoE fine-tuning at this sample
size. These results indicate that our sample size can statistically distinguish a single high-PFI cluster (MoE fine-tuning) from a continuum of lower-PFI
jobs.
Tables~\ref{tab:kw_groups} and~\ref{tab:dunn_pairs} provide group medians and pairwise comparisons, respectively.

\begin{table}[h]
  \centering
  \caption{PFI by task/architecture regime across the 25 headline sweep groups.}
  \label{tab:kw_groups}
  \begin{tabular}{lrrr}
    \toprule
    Group & $n$ & Median PFI & Mean PFI \\
    \midrule
    Dense (PT + LoRA SFT) & 14 & 1.27 & 1.30 \\
    MoE pretraining       &  7 & 1.43 & 1.42 \\
    MoE fine-tuning       &  4 & 2.10 & 2.17 \\
    \bottomrule
  \end{tabular}
\end{table}

\begin{table}[h]
  \centering
  \caption{Dunn's post-hoc test, all three pairs, Holm--Bonferroni corrected ($m = 3$).
  $^{*}$ denotes $p_{\mathrm{Holm}} < 0.05$.}
  \label{tab:dunn_pairs}
  \begin{tabular}{lcrrrc}
    \toprule
    Comparison & $n$ & $z$ & $p_{\mathrm{raw}}$ & $p_{\mathrm{Holm}}$ & Sig. \\
    \midrule
    Dense vs.\ MoE fine-tuning           & (14, 4) & $-3.44$ & 0.0006 & 0.0017 & $^{*}$ \\
    MoE pretraining vs.\ MoE fine-tuning & (7, 4)  & $-1.90$ & 0.0568 & 0.1137 &        \\
    Dense vs.\ MoE pretraining           & (14, 7) & $-1.64$ & 0.1020 & 0.1137 &        \\
    \bottomrule
  \end{tabular}
\end{table}

\subsection{Sensitivity analysis of fit PFI model}\label{app:dram_validation}

Table~\ref{tab:leverage_sensitivity} provides a sensitivity analysis for the DRAM copy product-based predictor, dropping high-leverage sweep groups and providing the Spearman correlation between DRAM copy product and PFI at each step.

Table~\ref{tab:dram_subgroup_rho} splits the correlation by architecture. We find no correlation within the dense subset, a positive but not significant correlation within the MoE subset ($\rho = 0.509$, $n=11$, $p = 0.11$), and a significant correlation over all groups ($\rho = 0.535$, $n=25$, $p = 0.006$).
As a result, we find that contrast between architecture-task clusters drives correlation between PFI and GPU telemetry, rather than variation within clusters.

\begin{table}[t]
  \centering
  \caption{Spearman rank correlation between the DRAM copy product and PFI as the $k$ highest-leverage sweep groups are removed, over the 25 headline groups. Leverage is computed from the predictor only. Groups are removed in leverage order: Qwen3-30B-A3B (MoE, fine-tuning, 8 GPU), gpt-oss-20b (MoE, fine-tuning, 8 GPU), Qwen3-30B-A3B (MoE, fine-tuning, 16 GPU), Qwen3-32B (dense, pretraining, 8 GPU), Llama-3.1-70B (dense, pretraining, 16 GPU).}
  \label{tab:leverage_sensitivity}
  \begin{tabular}{rrr}
    \toprule
    $k$ dropped & $n$ & $\rho$ \\
    \midrule
    0 & 25 & $0.535$ \\
    1 & 24 & $0.477$ \\
    2 & 23 & $0.406$ \\
    3 & 22 & $0.321$ \\
    4 & 21 & $0.438$ \\
    5 & 20 & $0.463$ \\
    \bottomrule
  \end{tabular}
\end{table}

\begin{table}[t]
  \centering
  \caption{Spearman rank correlation between the DRAM copy product and PFI within architecture subgroups, over the 25 headline sweep groups.}
  \label{tab:dram_subgroup_rho}
  \begin{tabular}{lrrr}
    \toprule
    Subgroup & $\rho$ & $n$ \\
    \midrule
    Dense only & $-0.178$ & 14 \\
    MoE only & $0.509$ &  11 \\
    All except MoE fine-tuning & $0.347$ & 21 \\
    All groups & $0.535$ & 25 \\
    \bottomrule
  \end{tabular}
\end{table}

\end{document}


\end{document}

%% file: intro.tex


Large language model (LLM) training is a rapidly growing source of data center electricity demand. Power availability has become a bottleneck for the deployment of new AI compute capacity, with limited grid capacity causing multi-year interconnection delays for new data centers~\cite{iea2026keyquestions}. Recent analysis of the U.S. power system found that data centers operating as flexible loads, reducing power consumption in a few peak hours, could unlock nearly 100 GW of additional capacity~\cite{norris2025rethinking}. The promise of such flexible capacity creates a strong incentive for AI training clusters to be power-flexible.


To realize these benefits, AI clusters must reduce power consumption when needed while maintaining acceptable training performance, as defined in customer service-level agreements (SLAs). Achieving SLA-aware power flexibility requires understanding the throughput cost of power reductions.


Prior studies of training efficiency primarily focus on energy consumption, Model FLOPs Utilization (MFU~\cite{chowdhery2023palm}), tokens-per-joule at nominal operating points~\cite{chung2025mlenergy}, or energy-delay tradeoffs for individual jobs~\cite{you2023zeus,chung2024perseus}. However, these metrics provide limited insight into the quantity most relevant for infrastructure control: a job’s throughput sensitivity to sustained reductions in available power for up to several hours during periods of peak grid load.

Understanding job-level power elasticity is important because training jobs with similar baseline throughput and power draw can respond very differently to the same GPU power cap. 
As a result, heuristics for spreading power constraints across jobs can waste throughput that could otherwise be preserved through elasticity-aware power allocation. To enable these strategies, cluster power managers must be able to identify which jobs can absorb power reductions at the lowest performance cost.


To this end, we present the first large-scale empirical characterization of the power elasticity of modern LLM training jobs 
under controlled GPU power modulation. We conduct 131 training runs across multiple architectures, pretraining and fine-tuning tasks, and up to 32 NVIDIA H200 GPUs, sweeping GPU power caps across the full operating envelope. 
From these measurements, we compute the Power Flexibility Index (PFI), a normalized metric that quantifies the performance cost of power reductions
%
and allows the cluster power manager to rank jobs by relative flexibility.

Building on this empirical characterization,  we develop telemetry-driven estimators that infer a job’s PFI online from readily-available signals from NVIDIA Data Center GPU Manager (DCGM), allowing cluster power managers to estimate PFI without power-cap sweeps.
We apply this estimator to simulated workloads to demonstrate how PFI-aware orchestration can protect relatively inflexible jobs while assigning deeper power reductions to relatively flexible jobs, maximizing tokens/sec under a global power budget.
%
%
Our contributions are:

\begin{itemize}
\item We present the first systematic characterization of \textbf{power elasticity} across LLM training jobs, across a range of model architectures, training tasks, and distributed GPU scales.
\item We introduce the \textbf{Power Flexibility Index (PFI)}, a normalized job-level metric that quantifies the throughput cost of power reduction.
\item We identify \textbf{runtime telemetry signals that predict PFI}, showing that elasticity can be inferred online from GPU monitoring signals that correlate with different flexibility regimes.
\item We demonstrate that \textbf{PFI-aware power allocation improves aggregate cluster throughput} under cluster-wide GPU power constraints, establishing a practical path toward power-aware, grid-responsive data center operation.
\end{itemize}

\begin{figure}[t]
  \centering
  \begin{minipage}[c]{0.23\linewidth}
    \includegraphics[width=\linewidth]{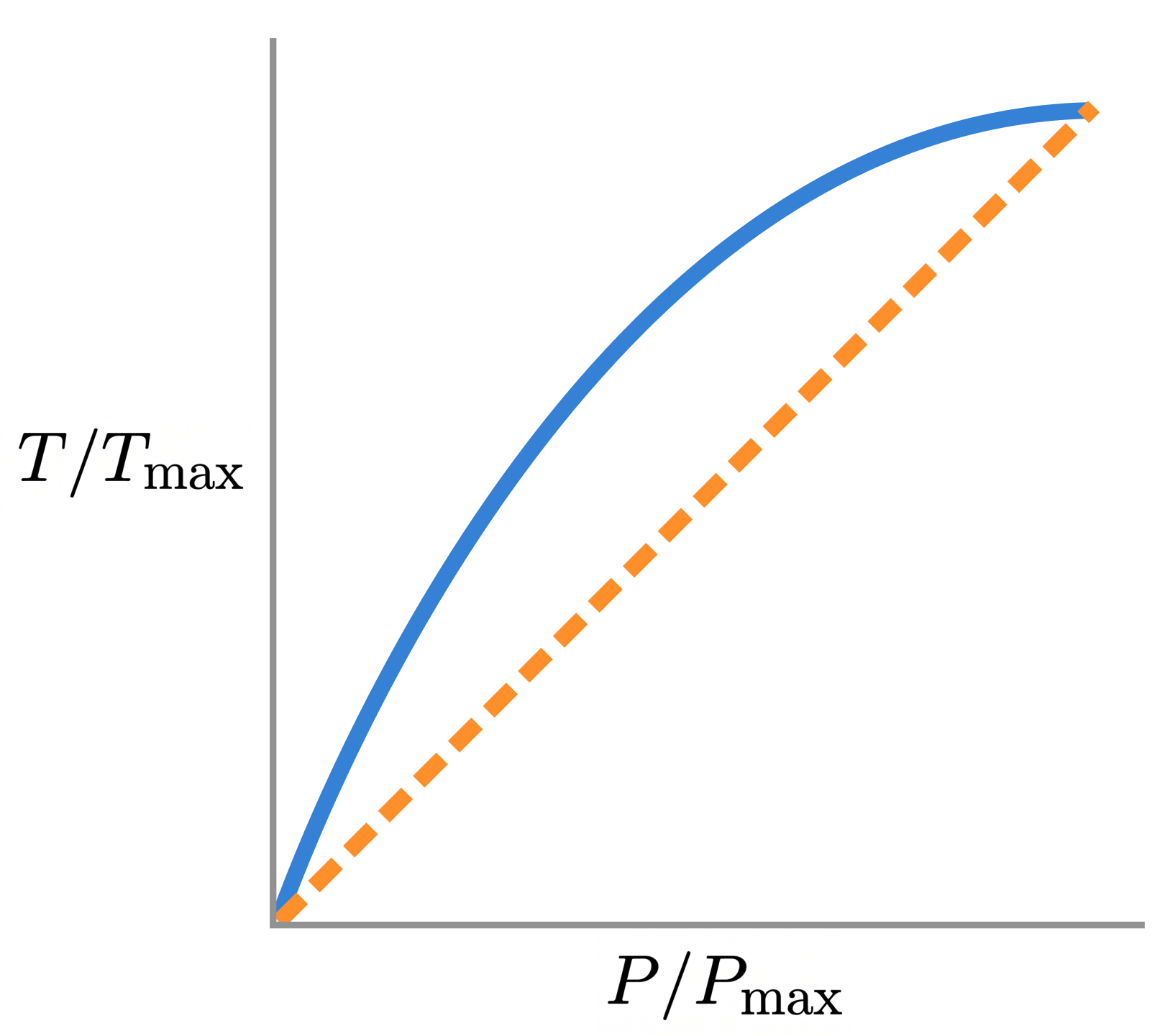}
  \end{minipage}%
  \hspace{0.5em}$\boldsymbol{\rightarrow}$\hspace{0.5em}%
  \begin{minipage}[c]{0.25\linewidth}
    \includegraphics[width=\linewidth]{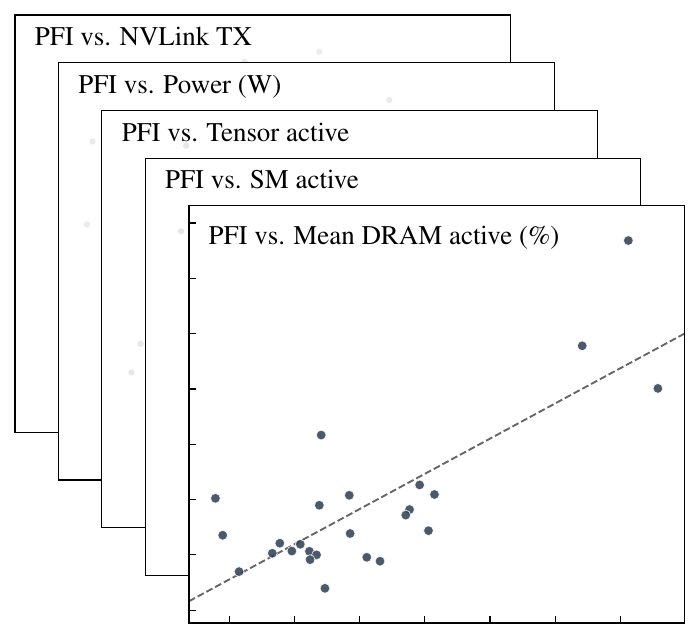}
  \end{minipage}%
  \hspace{0.5em}$\boldsymbol{\rightarrow}$\hspace{0.5em}%
  \begin{minipage}[c]{0.41\linewidth}
    \includegraphics[width=\linewidth]{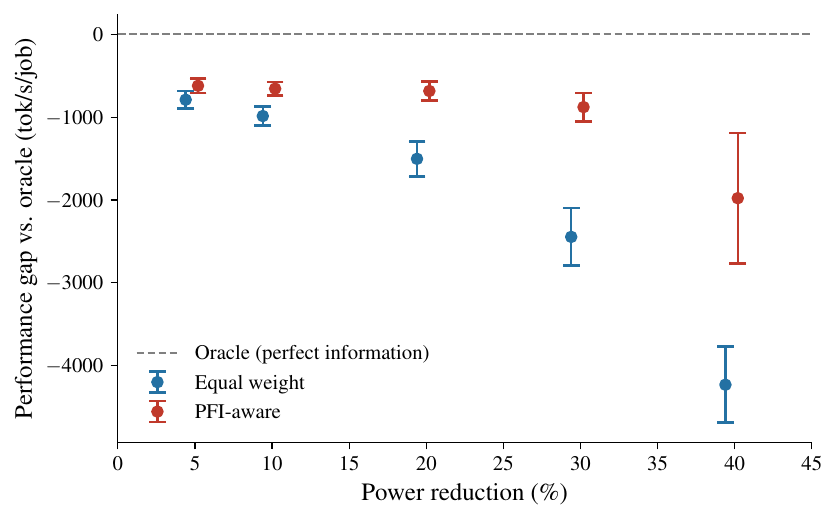}
  \end{minipage}
  \caption{\small{LLM training jobs vary in how quickly performance degrades under power reductions. Relatively \textcolor{captionblue}{more power flexible} ({\color{lineblue} ---}) jobs maintain throughput despite power reductions, while relatively \textcolor{captionorange}{less flexible} ({\color{lineorange} -\,-}) jobs quickly lose performance at lower power. We introduce the \textbf{Power Flexibility Index (PFI)}, a new metric for measuring power elasticity, and conduct large-scale, systematic characterization of modern LLM training jobs. We develop a model for predicting PFI online from readily available GPU telemetry, and we demonstrate how PFI-aware power allocation improves total throughput under cluster-wide GPU power constraints and realistic job mixes.}}
  \label{fig:hero}
\end{figure}

%% file: relatedwork.tex
\textbf{Hardware-level power management.}
GPU power capping and DVFS have been studied as efficiency knobs for over a decade. Tang et al.~\cite{tang2019gpudvfs} sweep core and memory frequency on Pascal and Volta GPUs for convolutional networks and identify workload-specific optimal frequencies. Krzywaniak et al.~\cite{krzywaniak2022gpucap} apply NVML power caps during CNN training on a single V100 or A100 and report 22 to 32\% energy savings at small slowdowns.
Zhao et al.~\cite{zhao2024sustainable} demonstrate that capping V100 power across a mixed HPC and AI workload can be done with minimal overall performance loss, and Costa et al.~\cite{toledocosta2025exascale} provide similar results on an exascale GPU system. For LLM workloads, Patel et al.~\cite{patel2024llmpower} characterize A100 cluster power for both training and inference and show the ability to increase capacity via power-capping-based oversubscription. Ujeniya et al.~\cite{ujeniya2026h100h200} compare power capping on H100 and H200 and show that memory bandwidth differences affect energy and performance tradeoff under power caps. The HPC community has explored related ideas on CPU-based systems~\cite{patki2015practical}. Prior work establishes power capping as a viable control knob but does not provide a job-level characterization of the power-performance tradeoff across architectures, cluster scales, and modern LLM training tasks.

\textbf{ML systems optimization.}
A large body of systems work studies ML throughput or energy optimization under fixed power budgets. The dominant building blocks include tensor and pipeline parallelism (Megatron-LM~\cite{shoeybi2019megatron}), sharded optimizer states (ZeRO~\cite{rajbhandari2020zero} and FSDP~\cite{zhao2023fsdp}), IO-aware kernels such as FlashAttention~\cite{dao2022flashattention}, and MoE training stacks~\cite{lepikhin2021gshard, fedus2022switch, hwang2023tutel, gale2023megablocks}. Complementary works such as Zeus~\cite{you2023zeus} and Perseus~\cite{chung2024perseus} optimize energy--time tradeoffs for training jobs, while ML.ENERGY~\cite{chung2025mlenergy} benchmarks inference energy use and recommends energy-efficient deployment configurations. While these works provide tools for optimizing energy use, our goal in this paper is to provide a characterization of job-level power flexibility that allows a training cluster to dynamically reduce power demand during periods of grid stress (rather than finding an efficient static operating point).


\textbf{Energy-aware and carbon-aware AI orchestration.}
A related line of work seeks to reduce the energy or carbon footprint of large-scale computing through job scheduling and control.
Carbon-aware computing frameworks shift flexible workloads in time or location in response to real-time electricity carbon intensity and power availability~\cite{radovanovic2021carbonaware, anderson2022treehouse}. DynamoLLM dynamically adjusts instance counts, routing policies, and GPU frequencies to minimize energy and carbon cost while satisfying latency SLAs in LLM inference clusters~\cite{stojkovic2025dynamollm}. By providing a systematic characterization of the power-throughput response of individual training jobs, our work complements these scheduler-level approaches to workload flexibility. 


\textbf{Demand response and power-flexible AI infrastructure.}
Data center demand response shifts computing demand in response to grid conditions~\cite{wierman2014datacentre,liu2011greening,zhangTPDS2022} or oversubscribes infrastructure through dynamic power provisioning~\cite{pelley2010power}. POLCA~\cite{patel2024polca} oversubscribes inference clusters using statistical headroom, while Wang et al.~\cite{wang2025reshaping} shape AI cluster power profiles for grid response by varying GPU frequency. At the operator and grid layer, field demonstrations~\cite{colangelo2025grid,williams2026powerflex}, together with EPRI's DCFlex initiative~\cite{eprimosaic}, have shown that GPU clusters can function as dispatchable grid resources. As with the energy-aware orchestration methods discussed above, our work complements these approaches by characterizing the job-level power-throughput response, which enables grid-responsive operation with minimal throughput loss.


\textbf{Positioning of this work.}
This paper fills an important gap by quantifying the throughput cost of power reductions for specific LLM training jobs. By introducing PFI and presenting the first (to our knowledge) systematic, large-scale characterization of power flexibility in LLM training, our work helps enable power-flexible data center operations by allowing schedulers to optimize power reductions to preserve cluster performance and maintain SLAs. In addition, by demonstrating that we can predict PFI from readily-available DCGM telemetry, we show that our approach is practically implementable and can measurably improve cluster performance on simulated workloads.
